\documentclass[letterpaper, 10pt, conference]{ieeeconf}

\IEEEoverridecommandlockouts

\usepackage{graphicx}
\usepackage{amsmath}
\usepackage{booktabs}
\usepackage{tabularx}
\usepackage{xspace}
\usepackage{pgfplots}
\pgfplotsset{compat=1.18}
\usepackage{float}
\usepackage{cite}
\usepackage{multirow}
\usepackage{siunitx}

\begin{document}

\title{\LARGE \bf ReGlove: A Soft Pneumatic Glove for Activities of Daily Living Assistance via Wrist-Mounted Vision}

\author{
    Rosh Ho$^{1}$ and Jian Zhang$^{1}$%
    \thanks{*This work was not supported by any organization}
    \thanks{$^{1}$Rosh Ho and Jian Zhang are students with the Department of Computer Science, Columbia University, New York, NY, USA. E-mail: \{r.ho, jz3607\}@columbia.edu}%
    \thanks{*Corresponding author: Rosh Ho, E-mail: r.ho@columbia.edu}%
}

\maketitle
\thispagestyle{empty}
\pagestyle{empty}

\begin{abstract}
This paper presents ReGlove, a system that converts low-cost commercial pneumatic rehabilitation gloves into vision-guided assistive orthoses. Chronic upper-limb impairment affects millions worldwide, yet existing assistive technologies remain prohibitively expensive or rely on unreliable biological signals. Our platform integrates a wrist-mounted camera with an edge-computing inference engine (Raspberry Pi 5) to enable context-aware grasping without requiring reliable muscle signals. By adapting real-time You-Only-Look-Once (YOLO) based computer vision models, the system achieves \SI{96.73}{\percent} grasp classification accuracy with sub-\SI{40.00}{\milli\second} end-to-end latency. Physical validation using standardized benchmarks shows \SI{82.71}{\percent} success on Yale-CMU-Berkeley (YCB) Object and Model Set object manipulation and reliable performance across \SI{27.00}{} Activities of Daily Living (ADL) tasks. With a total cost under \$\SI{250.00}{} and exclusively commercial components, ReGlove provides a technical foundation for accessible, vision-based upper-limb assistance that could benefit populations excluded from traditional Electromyography (EMG)-controlled devices.
\end{abstract}

\section{INTRODUCTION}
\label{sec:intro}

Upper-limb impairment resulting from stroke, spinal cord injury, or neuromuscular disorders affects over \SI{5.00}{million} Americans, significantly impacting independence and quality of life. While sophisticated robotic orthoses exist commercially, their high cost (often exceeding \$\SI{10000.00}{}) and complexity limit widespread adoption, particularly for chronic conditions requiring long-term use.

This work explores an alternative paradigm: functionally enhancing mass-produced, low-cost pneumatic rehabilitation gloves with vision-based control to create accessible assistive devices. Commercial pneumatic gloves present an attractive starting point, costing under \$\SI{50.00}{} while offering inherent compliance and safety through soft actuation. However, they typically operate through simple manual controls or require reliable surface electromyography (sEMG) signals—a significant limitation for patients with weak or noisy muscle activation due to neurological damage.

Recent advances in computer vision for prosthetic control demonstrate that visual context can robustly inform grasp selection \cite{degol2016, taverne2019}. However, these approaches have not been systematically applied to orthotic applications using commercial components. The ReGlove system bridges this gap by integrating established computer vision techniques with affordable, commercially available hardware.

This paper presents three key contributions: (1) An integrated hardware-software architecture that transforms commercial pneumatic gloves into vision-guided orthoses using readily available components; (2) A lightweight perception pipeline based on YOLO architectures that achieves real-time grasp classification on edge computing hardware; and (3) A comprehensive performance evaluation establishing baseline functionality across standardized benchmarks including YCB object manipulation and Activities of Daily Living (ADL) tasks. Through this proof-of-concept, we demonstrate a viable pathway toward assistive devices that balance capability with accessibility.

\section{RELATED WORK}
\label{sec:background}

\subsection{Actuation for Hand Assistance}
Hand assistive devices primarily employ cable-driven or pneumatic actuation. Cable-driven systems \cite{kim2025, rose2019} transmit force from proximal motors through tendon-like mechanisms, offering precise control but suffering from mechanical complexity, cable management issues, and limited compliance. 

Pneumatic actuators, used in commercial rehabilitation gloves, provide inherent compliance and safety through soft, inflatable chambers \cite{lim2023}. Clinical evidence supports their efficacy in improving hand function, with randomized trials showing significant improvements in active range of motion and grip strength for chronic stroke patients \cite{fardipour2022, ko2023}. Their commercial availability and low cost (\textless\$\SI{50.00}{}) make them a practical foundation for accessible assistive technology.

Alternative approaches include shape-memory alloys \cite{butzer2021} and motorized exoskeletons, but these face challenges in reliability, weight, and cost that limit practical deployment.

\subsection{Control Modalities}
Traditional control methods include manual triggers and sEMG. Manual control requires use of the contralateral limb, making it impractical for independent use. sEMG-based control can enable more natural actuation but often fails for patients with weak or noisy signals due to neuromuscular degeneration \cite{xu2025}.

Vision-based control, successfully demonstrated in prosthetic systems \cite{degol2016, zhang2025}, offers a promising alternative by relying on object context rather than biological signals. Prior work primarily used computationally intensive architectures like VGG-16, limiting real-time performance on low-power hardware. We adapt this approach using modern YOLO architectures optimized for edge deployment, making vision-based control practical for orthotic applications where EMG may be unreliable.

\section{SYSTEM DESIGN}
\label{sec:method}

\begin{figure}[htbp]
    \centering
    \includegraphics[width=0.8\columnwidth]{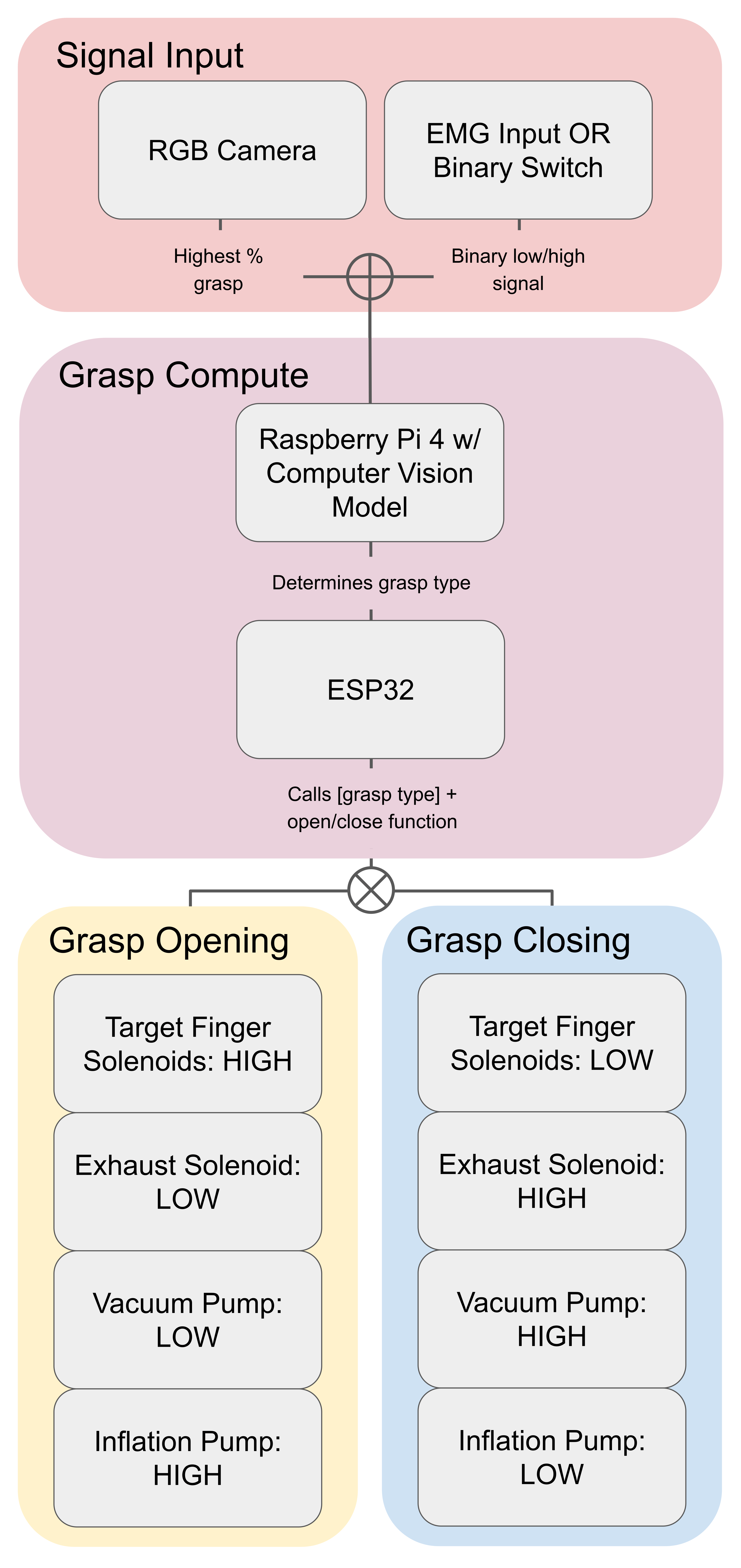}
    \caption{
        End-to-end system workflow.
        The wrist-mounted camera captures the visual scene and streams RGB frames to the Raspberry~Pi~5 for inference using the lightweight YOLO-based grasp classifier. 
        The predicted grasp type is forwarded to the ESP32 microcontroller, which manages valve-switching logic for the pneumatic circuit and actuates the glove accordingly. 
        A binary intent signal (tactile switch or sEMG) initiates the control loop, while the pumps and solenoid manifold generate positive or negative pressure to drive finger extension or flexion. 
        This diagram summarizes the integration of sensing, inference, pneumatic routing, and actuation within the complete assistive architecture.
    }
    \label{fig:workflow}
\end{figure}

The ReGlove system integrates a pneumatic glove with a vision-based control pipeline (Fig. \ref{fig:workflow}). A wrist-mounted camera captures the visual scene, a Raspberry Pi 5 runs the grasp classifier, and an ESP32 microcontroller operates the pneumatic components. A binary intent signal (tactile switch or sEMG) initiates the control loop.

\subsection{Hardware Implementation}
\label{subsec:hardware}

\begin{figure}[htbp]
    \centering
    \includegraphics[width=0.85\columnwidth]{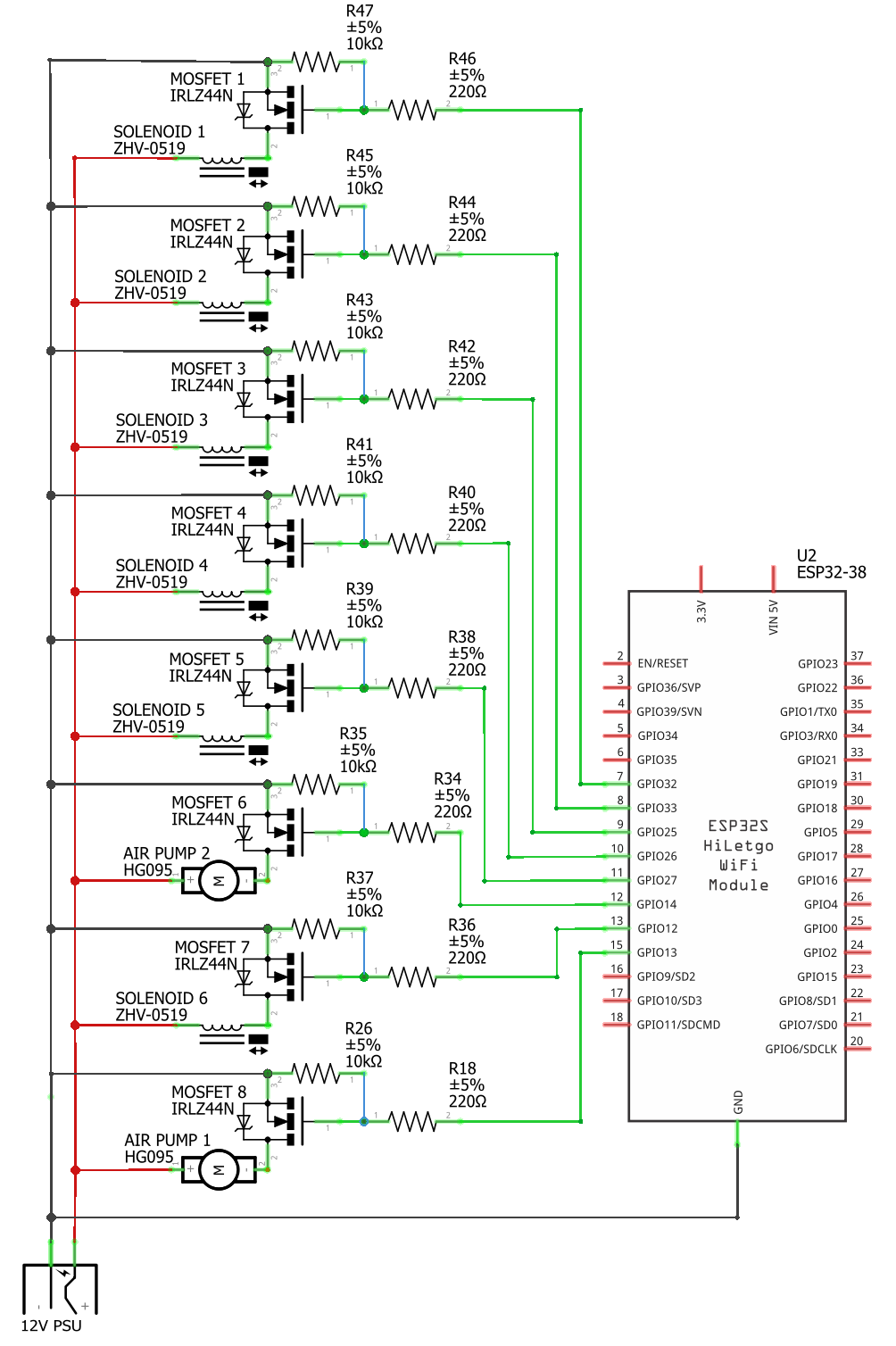}
    \caption{Complete wiring schematic for the pneumatic control system, illustrating connections between the Raspberry Pi 5, ESP32 microcontroller, solenoid valves, air pumps, and power supply components. The diagram shows both digital control signals and pneumatic pathways.}
    \label{fig:schematic}
\end{figure}

The pneumatic subsystem uses a commercial rehabilitation glove with ethylene-vinyl acetate bellows actuators, providing one degree of freedom per finger for bidirectional flexion and extension. We employ two HG095 mini air pumps (\SI{6.00}{\litre\per\minute} flow rate) for inflation and vacuum generation, and six ZHV-0519 three-way solenoid valves for individual finger control. The complete wiring schematic is shown in Fig. \ref{fig:schematic}. 

Safety Considerations: The system incorporates multiple safety features including an exhaust solenoid that actively regulates pressure during flexion cycles, preventing over-pressurization and ensuring fail-safe operation. This design eliminates risk of actuator failure or user injury from excessive pressure buildup, maintaining compliance with soft robotic safety standards for human-worn devices.

\begin{figure}[htbp]
    \centering
    \includegraphics[width=0.8\columnwidth]{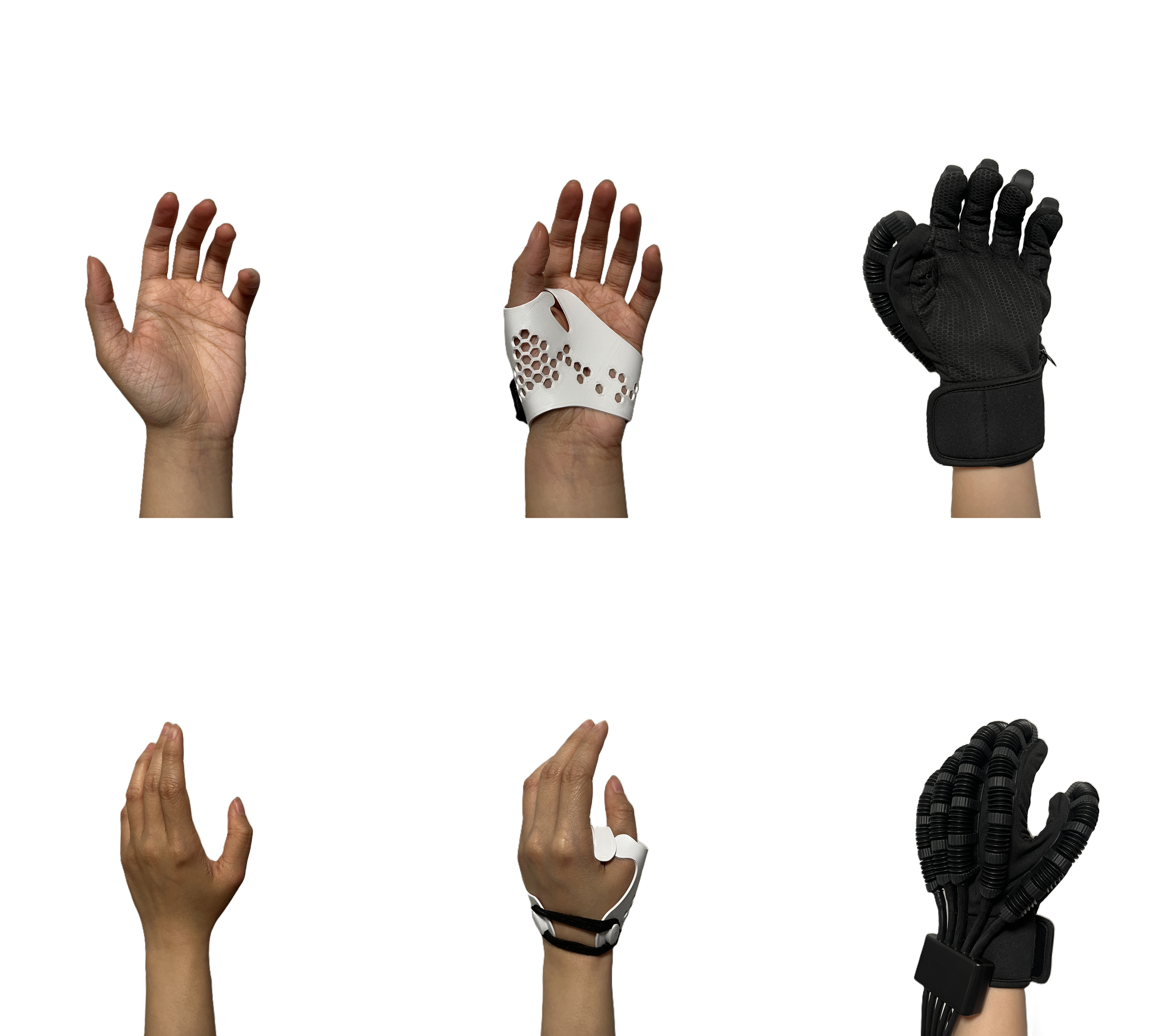}
    \caption{Hand configuration comparisons: (a) bare hand, (b) hand with 3D printed thumb brace, (c) complete orthosis glove worn over thumb brace. The brace maintains functional thumb positioning while allowing pneumatic flexion.}
    \label{fig:hand-glove-brace}
\end{figure}

Thumb Adaptation: The commercial glove's single degrees of freedom design limits thumb opposition. We address this with a custom 3D-printed thermoplastic polyurethane brace that maintains partial abduction while allowing pneumatic flexion, preserving capability for most functional grasp types \cite{zheng2011}. The hand configuration with and without the brace and glove is shown in Fig. \ref{fig:hand-glove-brace}.

Pneumatic Circuit: The system employs a semi-closed loop design with separate inflation and deflation subloops. During extension, the inflation pump activates while selected finger solenoids open; during flexion, the vacuum pump activates with reversed valve states. An exhaust solenoid regulates pressure between cycles.

Control Inputs: While the system architecture supports multiple input modalities (surface Electromyography, Electroencephalography, Electrooculography), we use a simple tactile switch for benchtop validation to isolate vision system performance. This allows future drop-in replacement with sEMG once IRB approval is secured for clinical studies.

The total hardware cost is approximately \$\SI{235.00}{} (Table \ref{tab:cost_breakdown}), with detailed specifications in supplementary materials.

\begin{table}[htbp]
\centering
\caption{Hardware Cost Breakdown (as of October 2025)}
\label{tab:cost_breakdown}
\begin{tabularx}{\columnwidth}{>{\raggedright\arraybackslash}X r}
\toprule
\textbf{Component} & \textbf{Cost (USD)} \\
\midrule
Pneumatic glove with finger control & \$17.00 \\
ZHV-0519 three-way solenoid valves (×6) & \$19.50 \\
Vinyl tubing (4 × \SI{5}{\milli\meter}) & \$7.50 \\
HG095 12 V DC, \SI{6}{\liter\per\minute} air pumps (×2) & \$3.46 \\
ESP32-WROOM-32D Microcontroller & \$4.29 \\
Raspberry Pi 5 (8 GB) & \$81.19 \\
Logitech c270 (wrist-mounted camera) & \$24.00 \\
MyoWare sEMG sensors & \$39.90 \\
IRLZ44N MOSFET (×8) & \$8.96 \\
12 V rechargeable battery & \$28.99 \\
\midrule
\textbf{Total} & \textbf{\$234.79} \\
\bottomrule
\end{tabularx}
\vspace{0.3em}
\end{table}

\subsection{Vision Pipeline \& Model Development}
\label{subsec:software}

We used a grasp classification system using three publicly available datasets: DeepGrasping (\SI{885.00}{} images) \cite{degol2016}, ImageNet subset (\SI{5180}{} images), and HandCam (\SI{250}{} images) \cite{taverne2019}. To address class imbalance, we applied extensive data augmentation including geometric transformations, photometric adjustments, and occlusion modeling, yielding approximately \SI{2000}{} images per grasp type (pinch, power, three-jaw chuck, tool, key).

We evaluated multiple architectures under identical training conditions:
\begin{itemize}
    \item VGG-16 \& VGG-16 + Depth: Baseline models replicating prior work \cite{degol2016}
    \item YOLO v11 \& v12: Modern lightweight object detectors optimized for edge deployment
\end{itemize}

Depth augmentation using synthetic depth maps from DepthAnything \cite{yang2024} did not improve performance, likely due to inconsistency in synthetic depth quality. Both YOLO variants significantly outperformed VGG-based approaches (Table \ref{tab:model_comparison}), with YOLO v11 achieving \SI{96.67}{\percent} accuracy versus \SI{82.59}{\percent} for VGG-16. YOLO's superior performance stems from architectural features that preserve spatial structure (Spatial Pyramid Pooling -
Fast, Feature Pyramid Network/Path Aggregation Network layers) and integrated augmentation mechanisms that improve robustness to lighting and background variation.

Given its optimal accuracy-latency tradeoff, we selected YOLO v11 for system integration, achieving \SI{0.90}{\milli\second} inference latency on Raspberry Pi 5—well below the \SIrange{10.00}{20.00}{\milli\second} threshold for human-perceptible feedback \cite{jerald2009relatingscene}.

\begin{table}[htbp]
\centering
\caption{Grasp Classification Model Performance Comparison}
\label{tab:model_comparison}
\begin{tabularx}{\columnwidth}{l c @{\extracolsep{\fill}} r}
\toprule
\textbf{Model} & \textbf{Accuracy (\%)} & \textbf{Inference Time (ms)} \\
\midrule
VGG-16 & \SI{82.59}{} & \SI{7.24}{} ± \SI{0.45}{} \\
VGG-16 + Depth & \SI{79.91}{} & \SI{7.32}{} ± \SI{0.52}{} \\
YOLO v11 & \SI{96.67}{} & \SI{0.90}{} ± \SI{0.15}{} \\
YOLO v12 & \SI{96.45}{} & \SI{0.50}{} ± \SI{0.08}{} \\
\bottomrule
\end{tabularx}
\end{table}

\section{EXPERIMENTAL RESULTS}
\label{sec:results}

\subsection{Grasp Classification Performance}
The YOLO v11 model achieved a mean grasp classification accuracy of \textbf{\SI{96.67}{\percent}} (\SI{95.00}{\percent} CI: \SIrange{95.20}{97.80}{\percent}) on the test set. Analysis of the confusion matrix (supplementary Fig. S1) revealed that most misclassifications occurred between geometrically similar pinch and three-jaw chuck grasps. Performance degradation was primarily observed for scale-ambiguous objects where visual cues alone were insufficient to infer absolute size.

The model's inference latency of $\SI{0.90}{} \pm \SI{0.15}{}$ ms enables real-time operation, with total image preprocessing and classification completing in under \SI{2.00}{ms}. This represents a \SI{8.00}{\times} speedup compared to VGG-16 while maintaining superior accuracy.

\subsection{Physical Grasping Performance}
We evaluated physical grasping capability using standardized benchmarks to assess functional utility.

\subsubsection{YCB Object Set}
Using the YCB Gripper Assessment Protocol \cite{calli2015}, ReGlove achieved an overall success rate of \textbf{\SI{82.71}{\percent}} (\SI{215.50}{}/\SI{260.50}{} points). Performance was robust for objects with defined edges and surfaces (cups, blocks, utensils) but lower for small, smooth, or low-friction items (marbles, coins, washers). This performance gap primarily reflects mechanical limitations of the compliant ethylene-vinyl acetate actuators rather than perception errors. Full results are available in supplementary materials (Table S-III).

\subsubsection{Activities of Daily Living (ADL)}
\begin{figure*}[htbp]
\centering
\begin{tikzpicture}
\begin{axis}[
  xbar,
  xmin=0,
  width=1.7\columnwidth,
  height=0.7\textheight,
  xlabel={Execution Time (s)},
  ylabel={ADL Tasks},
  ylabel style={at={(-0.15,0.5)},anchor=south}, 
  symbolic y coords={Strap shoes, Open handbag, Don glasses, Comb hair, Apply toothpaste, Unpack tablets, Glue, Paint, Screw a bolt, Erase, Write, Open pen, Fold shirt, Iron, Grasp garment, Fluff up a pillow, Sweep dust, Sweep the floor, Serve wine, Pour wine, Pour juice, Serve food, Put skillet on stove, Open jar, Pour coffee, Fill in coffee, Fill water},
  ytick=data,
  bar width=5pt,
  enlarge y limits=0.02,
  legend style={
    at={(0.5,-0.075)}, 
    anchor=north, 
    legend columns=2,
    /tikz/every even column/.append style={column sep=0.5cm},
    fill=none
  },
    legend image code/.code={
    \draw[#1] (0cm,-0.1cm) rectangle (0.3cm,0.1cm);
  },
  nodes near coords,
  nodes near coords style={font=\tiny},
  point meta=explicit symbolic,
  every node near coord/.append style={xshift=3pt},
  grid=major,
  grid style={dashed,gray!30}
]

\addplot+[
  fill=blue!30,
  draw=blue,
  bar shift=-2.5pt
] table [
  x=Avg Human Execution Time (s),
  y=Tasks,
  meta=Avg Human Execution Time (s),
  col sep=comma
] {datasets/ADL.csv};
\addlegendentry{Human Execution Time}

\addplot+[
  fill=red!30,
  draw=red,
  bar shift=2.5pt
] table [
  x=Avg ReGlove Execution Time (s),
  y=Tasks,
  meta=Avg ReGlove Execution Time (s),
  col sep=comma
] {datasets/ADL.csv};
\addlegendentry{ReGlove Execution Time}

\end{axis}
\end{tikzpicture}
\caption{Comparative analysis of human versus ReGlove execution times across \SI{27.00}{} daily living tasks tasks. Blue bars represent average human performance, while red bars show ReGlove-assisted performance.}
\label{fig:adl_performance}
\end{figure*}
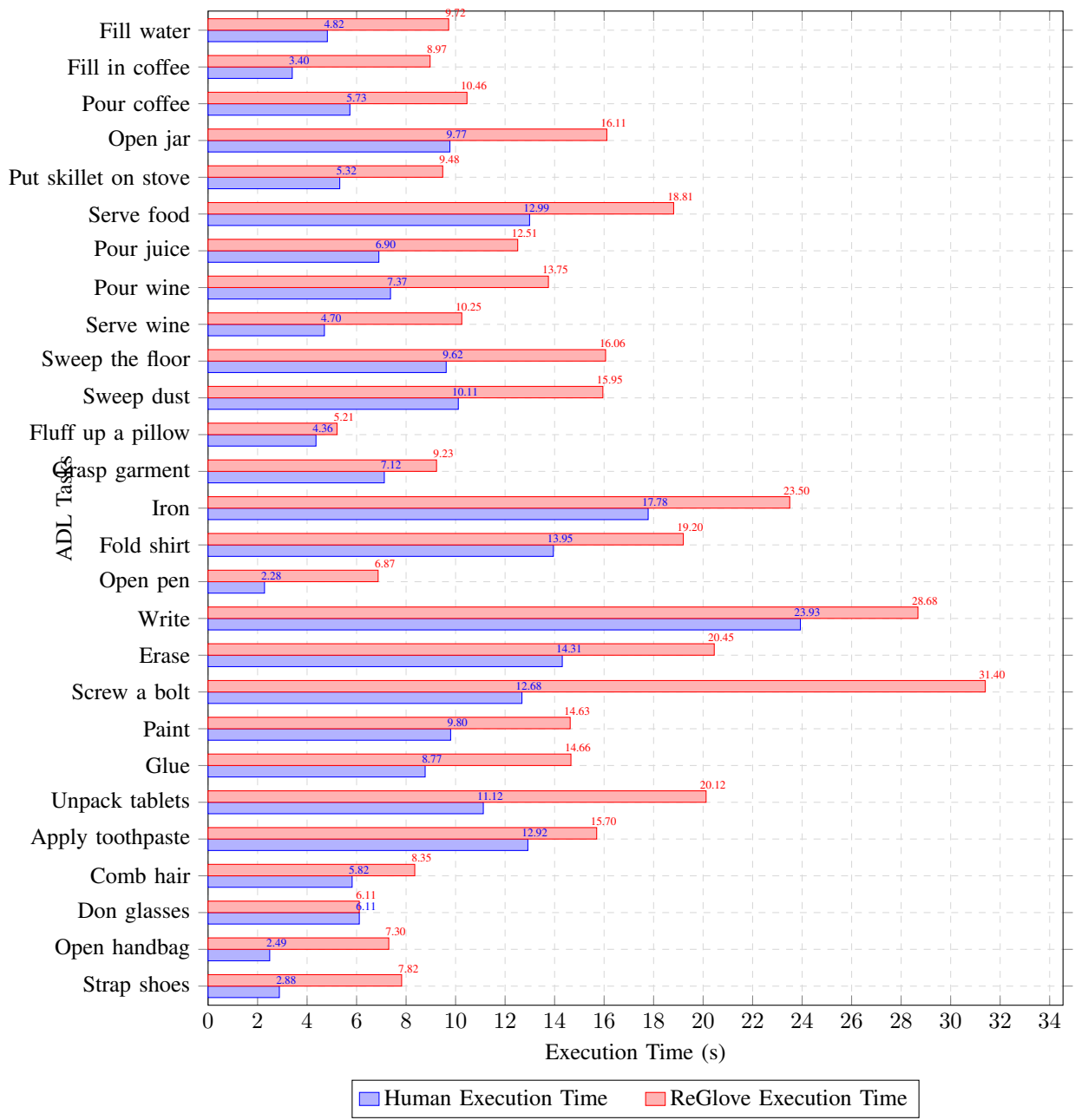

On a subset of \SI{27.00}{} daily living tasks based on Matheus \& Dollar \cite{matheus2010}, the system achieved a mean performance score of \textbf{$\SI{2.65}{} \pm \SI{0.28}{}$} out of \SI{3.00}{} (\SI{0.00}{}=failed, \SI{3.00}{}=excellent). The system excelled at tasks involving power or tripod grasps (pouring liquids, manipulating utensils) but struggled with fine manipulation requiring precise fingertip control (unwrapping tablets, rotating small bolts). A comparative analysis of execution times for these tasks is shown in Fig. \ref{fig:adl_performance}.

Multi-phase operations revealed limitations in sequential grasp switching, highlighting the need for more sophisticated control hierarchies. Complete task-by-task results are provided in supplementary materials (Table S-IV, Figure S2).

\subsection{Integrated System Performance}
The complete assistive system achieved end-to-end latency of $\SI{38.00}{} \pm \SI{6.40}{}$ ms from image capture to glove actuation, confirming real-time responsiveness for interactive use. The system reliably executed all five grasp types under live inference conditions without performance degradation during extended operation.

During \SI{90.00}{}-minute continuous testing sessions, the waist-mounted pneumatic unit maintained stable operation without overheating or pressure drift. Average power consumption was $\SI{10.30}{} \pm \SI{1.20}{}$ W, compatible with commercially available \SI{12.00}{V} portable battery packs for untethered operation.

\begin{table}[htbp]
\centering
\caption{Summary of System Performance Evaluation}
\label{tab:summary}
\begin{tabularx}{\columnwidth}{l @{\extracolsep{\fill}} r}
\toprule
\textbf{Metric} & \textbf{Performance} \\
\midrule
\textbf{Software Performance} & \\
\quad Grasp Classification Accuracy & \SI{96.67}{\percent} \\
\quad Inference Latency & $\SI{0.90}{} \pm \SI{0.15}{}$ ms \\
\addlinespace[0.5em]
\textbf{Hardware Performance} & \\
\quad YCB Object Success Rate & \SI{82.71}{\percent} \\
\quad ADL Task Score (\SIrange{0.00}{3.00}{}) & $\SI{2.65}{} \pm \SI{0.28}{}$ \\
\addlinespace[0.5em]
\textbf{Integrated System} & \\
\quad End-to-End Latency & $\SI{38.00}{} \pm \SI{6.40}{}$ ms \\
\quad Average Power Draw & $\SI{10.30}{} \pm \SI{1.20}{}$ W \\
\quad Continuous Operation Duration & \SI{90.00}{} minutes \\
\bottomrule
\end{tabularx}
\end{table}

\section{DISCUSSION}
\label{sec:discussion}

The ReGlove system demonstrates that commercial pneumatic rehabilitation gloves can be effectively converted into vision-guided assistive orthoses through integration with modern computer vision and low-cost computing hardware. This approach offers a affordable (under \$\SI{250.00}{}), non-invasive pathway toward functional hand assistance that circumvents the limitations of EMG-based control.

\subsection{Technical Performance and Significance}
The system's \SI{96.67}{\percent} grasp classification accuracy and \SI{38.00}{ms} end-to-end latency compare favorably with prior vision-based prosthetic systems requiring more complex hardware \cite{degol2016, taverne2019}. More significantly, by relying exclusively on visual context rather than biological signals, the approach extends accessibility to patient populations with unreliable EMG due to neuromuscular degeneration \cite{xu2025}.

The performance gap between software perception (\SI{96.67}{\percent} accuracy) and physical execution (\SI{82.71}{\percent} YCB success) highlights the mechanical limitations of commercial pneumatic gloves rather than perception shortcomings. This suggests that relatively simple hardware improvements—such as high-friction fingertip coatings or reinforced actuator segments—could significantly enhance functional performance without increasing system complexity or cost.

\subsection{Limitations and Design Considerations}
Several important limitations warrant discussion. The current "pause-and-select" control paradigm requires users to position their hand and trigger a single, static grasp. This does not support dynamic tasks requiring mid-manipulation grasp adjustments or provide mechanisms for user correction of mispredicted grasps.

The system's performance with small, smooth objects remains limited by the compliant nature of pneumatic actuation. While this compliance enhances safety, it reduces precision for fine manipulation tasks. Future iterations could incorporate variable-stiffness mechanisms or hybrid actuation approaches to balance safety and dexterity.

Our benchtop validation used a healthy operator, which allowed controlled testing of core functionality but leaves open questions about real-world performance with impaired users. The simplified binary intent detection (tactile switch) served as a reliable trigger for technical validation but may not reflect the control challenges faced by target users.

\subsection{Future Directions}
Building on this proof-of-concept, several research directions appear promising:

\begin{itemize}
\item Multi-modal control integration: Subsequent iterations will incorporate sEMG as the primary intent detection modality, operating in concert with the existing vision-based grasp classification. This hybrid approach will enable more natural actuation paradigms while maintaining the robustness of visual context awareness. Additionally, implementation of closed-loop force control will enhance manipulation precision and user experience.

\item Hardware refinement: Improved actuator geometry, high-friction surfaces, and variable-stiffness mechanisms to enhance grip stability and fine manipulation capability.

\item Control hierarchy expansion: Temporal grasp sequencing and gesture prediction to enable complex, multi-phase tasks like opening containers or using tools.

\item Clinical translation: Formal studies with stroke and SCI patients to quantify daily living tasks improvement, user acceptance, and long-term usability.

\item System integration: Miniaturization of pneumatic components and development of fully self-contained wearable form factors.
\end{itemize}

The modular architecture supports incremental improvement in each of these areas while maintaining the core benefits of affordability and accessibility.

\section{CONCLUSION}
\label{sec:conclusion}

This work presents ReGlove, an end-to-end demonstration of vision-guided pneumatic hand assistance using exclusively commercial components and open-source software. The system achieves real-time dexterous grasping with \SI{96.67}{\percent} classification accuracy and \SI{82.71}{\percent} physical success on standardized benchmarks, while maintaining a total cost under \$\SI{250.00}{}.

By bridging affordable rehabilitation hardware with modern computer vision, ReGlove offers a practical pathway toward restoring functional hand capability for individuals with chronic upper-limb impairment. The approach demonstrates that intelligent assistive technology need not be complex or expensive to be effective, providing a foundation for future development of accessible devices that can significantly impact quality of life for underserved populations.

\section*{FUNDING}
This research received no external grants or funding. It was supported by the authors' personal resources.

\section*{SUPPLEMENTARY MATERIALS}
\label{sec:supplementary}
Additional materials are available as ancillary files with this arXiv submission, including:
\begin{itemize}
    \item Confusion matrix analysis (Fig. S1)
    \item Complete YCB benchmark results (Table S-III)
    \item Detailed ADL task performance (Table S-IV, Figure S2)
    \item Hardware specifications and wiring diagrams
\end{itemize}

\bibliographystyle{IEEEtran}
\bibliography{references}

\end{document}